\documentclass[11pt,a4paper]{article}
\usepackage[hyperref]{acl2018}
\usepackage{times}
\usepackage{latexsym}
\usepackage{amsmath}
\usepackage{listings}
\usepackage{graphicx}
\usepackage[]{algorithm2e}

\usepackage{url}

\aclfinalcopy 

\title{syntree2vec: An algorithm to augment syntactic hierarchy into word embeddings}

\author{Shubham Bhardwaj \\
  VIT, Vellore, India \\
  {\tt shubham.bhardwaj4245@gmail.com}}

\date{1st March 2018}

\begin{document}
\maketitle
\begin{abstract}
Word embeddings aims to map sense of the words into a lower dimensional vector space in order to reason over them. Training embeddings on domain specific data helps express concepts more relevant to their use case but comes at a cost of accuracy when data is less. Our effort is to minimise this by infusing syntactic knowledge into the embeddings. We propose a graph based embedding algorithm inspired from node2vec \cite{grover2016node2vec}. Experimental results have shown that our algorithm improves the syntactic strength and gives robust performance on meagre data.
\end{abstract}

\section{Introduction}

Word embeddings are a set of continuous values vectors which are meant to embed the meaning of each word. These embeddings can then be used to perform semantic reasoning. The idea being words having same meaning are embedded closely in the vector space. It was first proposed by \citeauthor{bengio2003neural} but gained world wide acceptance with \citeauthor{mikolov2013distributed} seminal paper on word embeddings called Word2Vec. We take the Word2Vec model as starting point and address the shortcoming of their neighborhood sampling mechanism. Their neighborhood sampling mechanism ignores the word order which results in loss of some essential syntactic information.

Recently some attempts have been made in this direction such as PENN \cite{trask2015modeling} where the word embedding is partitioned and each of the partition is trained separately making them learn different senses of the same word. The partition is based on each word's relative position probability to the focus term. Their word sampling is directional in nature. On a similar line another algorithm was proposed by \cite{ling2015two} where they propose a structured skip-gram model. Here they do not provide any experiments with smaller size of data. Since we are working towards providing quality embeddings on small data we need to make sure that there is minimum loss of information. Therefore we explored the space of graph based algorithms because they tend to preserve structural properties when transforming graph into vector space \cite{cai2018comprehensive}. Recent work in graph based embeddings for text is based on co-occurrences of words, if the co-occurrence is higher then the strength of connection link between nodes(words) is higher in the graph. We are concatenating multiple dependency parse trees to create our graph instead. This ensures that a lot of rich information about syntax is not lost.

We propose syntree2vec, a graph embedding algorithm where we try to preserve the word order during sampling. The dataset chosen is the wikipedia articles archive 2008 \cite{schnabel2015evaluation}. We are using the syntaxnet model \cite{alberti2017syntaxnet} for dependency parsing. 
syntree2vec generates word embeddings by optimizing a graph based objective function that maximizes the probability of preserving a network neighborhood. We use a semi-guided second order random walk procedure to sample network neighborhood of nodes. Our method modifies the node2vec graph algorithm's flexible neighborhood objective to be specialized to work for creating word embeddings.  

The novelty of our work is 2 fold -
\begin{enumerate}
    \item The algorithm tends to preserve the word order during sampling
    \item The algorithm helps produce word embeddings that performs well on sequence generation task on less data.
\end{enumerate}

The success of a word embedding for a specialized task depends on how much of external information that has been encoded in it can be recovered from it \cite{jastrzebski2017evaluate}. In practice much of the pre-trained embeddings provide only little benefit under various settings \cite{zhang2015character} and so the embeddings should also be evaluated for variying data sizes given the rise of small data applications. Therefore we need to take into account the data efficiency of word embeddings also. To summarise our experiments will be based on the following 2 questions - 
\begin{enumerate}
\item How much syntactic hierarchy can we recover - this would suggest how much information loss occurred during the feature selection phase?
\item How does the performance vary with varying input size of data?
\end{enumerate}
For framing an experiment for question 1 we need to adapt to a transfer learning view of the process. It is a well known fact that Natural Language problems are very sensitive to problem domains, therefore, to transfer the knowledge of syntactic hierarchy we need to choose a simple supervised task which would benefit from this word embedding's specialty. For this task we have chosen word generation. In this task we try to predict the next word given the current sequence. The setting will have one word fed to a recurrent neural network and it will try to generate a sentence. The intuition being that if the knowledge of syntactic hierarchy is present, we would then get more syntactically relevant words generated. We would evaluate its performance and report the results on varying data sizes.

\section{Related Work}

There are 3 models which play a major role in inspiring this algorithm. Namely -
\subsection{Word2Vec}

It is a neural prediction model for representing continuous representation of words in low dimensional space. It is one of the most popular word embeddings available. It tries to maximise the likelihood of target word given history. The author of this model proposed architectures where the same objective is maximised in the other way around also i.e (skip-gram and CBOW model). Our algorithm improves upon their sampling procedure trading off some training time for better accuracy on the supervised task. The next model sets the horizon for outlining our research inspiration.

\subsection{DEPS}
DEPS\cite{levy2014dependency} tries to preserve the word order during the sampling by introducing words that are dependent to each other in a sentence to be part of a common neighborhood. For this he modified the skip gram model to incorporate variable length contexts. Dependency parsing was done using the Stanford Parser \cite{goldberg2012dynamic}. What we need here is a strategy that does the above job along with keeping a global view rather than being restricted sentence-wise which is the idea of the next model to be introduced. The next model showed such properties.

\subsection{Node2Vec}
Node2Vec is a graph embedding algorithm that works well over homogeneous networks. The advantage of using this algorithm is its flexible objective which gives a good balance of trade off between homophily and structure equivalence. This is a huge plus over DEPS, which has a more local neighborhood sampling criteria. The sampling strategy encourages them to be nearer together. This algorithm is the basis of our sampling strategy. What we would also like to  incorporate is a small bias towards sampling the next words based on how much do their tags occur together in order to make use of the syntactic information present in the graph. For example in the following sentences -
\begin{verbatim}
1. Ram(NOUN) works(ROOT) at apple
2. Sheela(NOUN) sleeps(ROOT)
   in the bedroom
\end{verbatim}
the NOUN and ROOT tags co-occur together often. Therefore during our sampling procedure if we encounter a noun as the focus word then the probability of taking a ROOT as the neighbor should be higher. This tag frequency count is maintained in what we would introduce in future subsection - the tag-transition matrix. 
Our algorithm exploits the richness of information in syntax trees and also incorporates the structural equivalence criteria in our algorithm.

\section{Algorithmic Framework}

We first introduce you to the general framework of a present day word-embedding model. We take the skip-gram objective as reference. Our aim is to get a vector representation of each word in the corpus into an n-dimensional vector space. Given a sequence of training words $w_1, w_2, w_3 ,.., w_N$ The objective is to try to maximise the average log probability-
\label{ssec:eqn}
\begin{equation}
1/N\sum_{n=1-c\leq j\leq c, j \neq 0}^{N}\sum log(p(w_{n+j} |w_n))
\end{equation}
where $p(w_{t+j} |w_t)$ is the probability of seeing the next word given the previous word and c is the training context. The weights of the neural network trained on this objective are then used as the corresponding word embedding for the input words used during its training. 

We try to address the task of creating embeddings using an alternate approach of graph embeddings. A graph embedding model tries to represent each node of a graph as a n-dimensional vector in a vector space. So our first task is to work out an efficient method to represent text in the form of a graph. If each word is a node then we are performing the task as a standard word embedding model. The graph creation algorithm is highlighted below.

\subsection{Graph creation}
\begin{figure}
  \includegraphics[width=\linewidth]{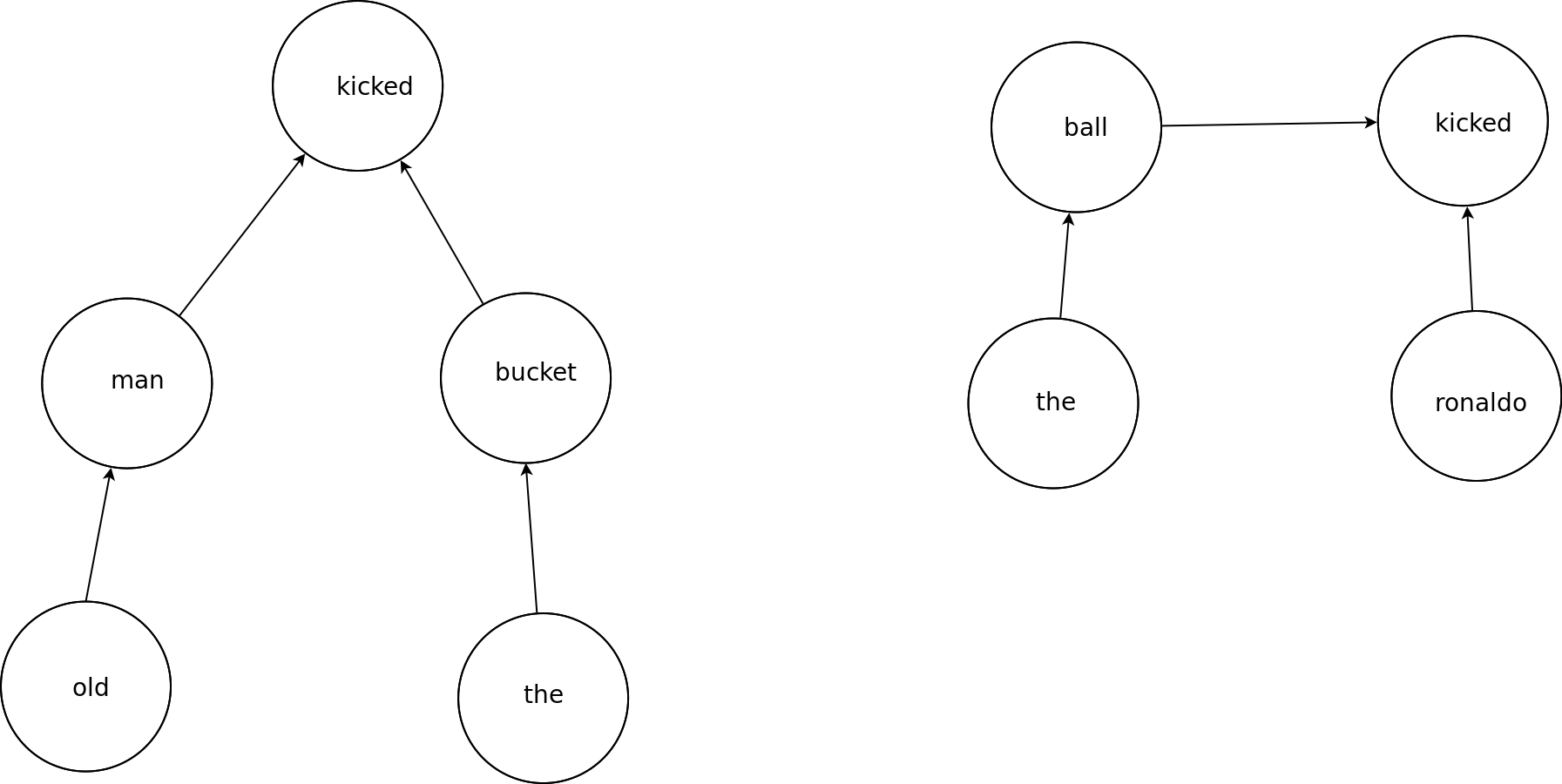}
  \caption{Individual parse trees of the 2 example sentences}
  \label{fig:graph_1}
\end{figure}

\begin{figure}
  \includegraphics[width=\linewidth]{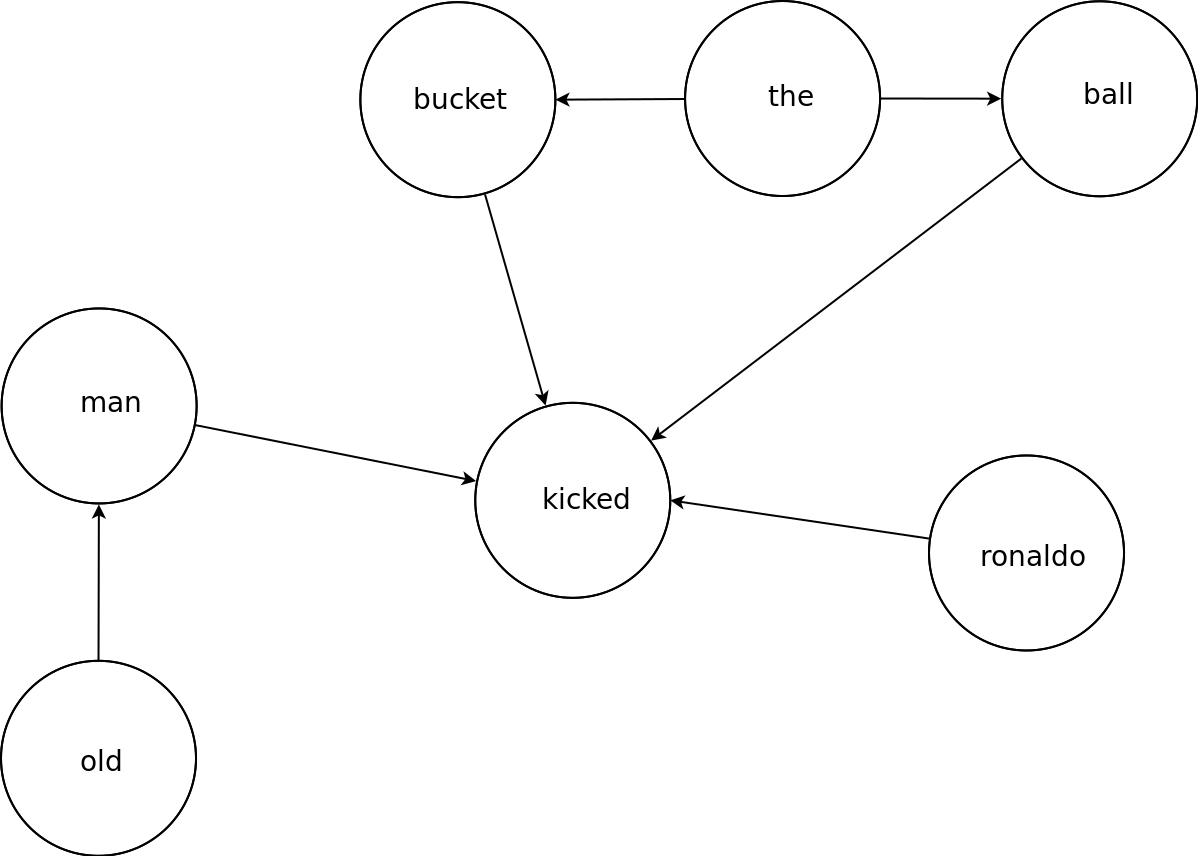}
  \caption{Resultant concatenated graph}
  \label{fig:giant_graph}
\end{figure}

The input comprises of a list of sentences. We feed each input sentence to syntaxnet dependency parser to get a parse tree. We concatenate the parse trees of all sentences on common nodes. Along with this we keep track of the relative tag co-occurrence with the help of the tag-transition/adjacency matrix. We illustrate the graph concatenation procedure with a small example given below.
Consider the 2 example sentences - 
\begin{verbatim}
1. The old man kicked the bucket
2. Ronaldo kicked the ball
\end{verbatim}
Their individual parse trees are shown in Figure \ref{fig:graph_1}.
The resultant concatenated graph a.k.a giant graph is shown in Figure  \ref{fig:giant_graph}

On the graph obtained from the above procedure we try perform a semi-supervised random walk whose objective is given below.
\subsection{Objective}
The objective that we are trying to optimize is same as the one given in node2vec presented in the equation below. It tries to maximize the log probability of seeing the network neighborhood N of node u given its feature vector f(u) and sampling strategy s.
\label{ssec:eqn}
\begin{equation}
max_f \sum_{u\epsilon V}\log Pr(N_s(u)|f(u))
\end{equation}
The notion of neighborhood $N_s(u)$ in skip-gram was defined using the sliding-window. We select the neighbors of the node u by performing a biased 2nd order random walk.

\subsection{Random Walks}
Random walk here refers to the event where we traverse through the nodes of a graph in a sequential random order. \cite{avin2008power}. Let $c_k$ be the $k^{th}$ node in the walk. With the starting node as u. The probability of visiting this node $c_k$ from the previous node $c_{k-1}$ is mathematically given as -

\label{ssec:eqn}

\begin{equation}
P(c_{k}, c_{k-1}) = 
\begin{cases}
\frac{{\pi_{vz}}}{Z}  &  if(v, x)  \in E \\
0   & otherwise
\end{cases}
\end{equation}
where $\pi_{vz}$ is the unnormalized transition probability between the nodes v,x and Z is the normalizing constant. The unnormalized transition probabilities also includes the addition of tag frequencies introduced by the tag transition matrix. The value of $\pi_{vz}$ calculated using the search bias.
\subsection{Search bias   \(\alpha\)}
Here the search bias is modified using the tag transition matrix such that the random walk is encouraged to move towards next neighbor based on the discrete normalized probability values for each of its focus word - neighbor corresponding tag frequency present in the matrix. This method in a way inhibits the movement of random walk from leaf nodes since there are less nodes connected to it, that lie at the end of parse trees, the tag transition matrix for them is sparse. For solving this problem we sample the nodes for them based on static edge weights.
we make use of the search parameters p and q which allow us to interpolate between BFS and DFS type search strategies. We can tune them to get a mix of such behaviour based on our application. The search bias \(\alpha\) is calculated as -  
\label{ssec:eqn}
\begin{equation}
\alpha_{pq}(t,x) = \
\begin{cases}
\frac{1}{p}  &  if  d_{t,x} = 0  \\
 1  &  if d_{t,x} = 1  \\
\frac{1}{q}  &  if  d_{t,x} = 2 
\end{cases}
\end{equation}

\subsection{The syntree2vec algorithm}
\begin{algorithm}
\DontPrintSemicolon 
 \caption{LearnFeatures()}

\KwIn{number of walks r, walk length l, context size k, return p, In-out q, dimensions d, Graph G = (V, E, W)}
\KwOut{an array of feature vectors f}
$\pi \gets PreprocessWeights(G, p, q, TTMat)$\;
$G \gets (V  , E, \pi)$\;
$walks \gets \{\}$\;

\For{$i \gets 1$ \textbf{to} $r$} {
  \ForEach{$node \in  V$} {
     $walk \gets syntree2vecWalk( G,  node,  l)$\;
    }
     $walks \gets walks \cup walk$\;
  
}
$f \gets StochasticGradientDescent(k, d, walks)$\;
\Return{f}\;

\label{algo:learn_features}
\end{algorithm}
\begin{algorithm}
\DontPrintSemicolon 
 \caption{syntree2vecWalk()}

\KwIn{start node u, walk length l, Graph G = (V, E, W)}
\KwOut{walk}
$walk \gets  \{u\}$\;

\For{$iter \gets 1$ \textbf{to} $l$} {
  $curr \gets walk[-1]$\;
  $ V_{curr} \gets GetNeighbors(curr, G) $\;
  $s \gets AliasSample(V_{curr} , \pi)$\;
  $walk \gets walk \cup s$\;

}

\Return{walk}\;

\label{algo:sytree_walk}
\end{algorithm}
The features provided as output correspond to representations learned for each word.
\section{Experiments}
\begin{figure}
  \includegraphics[width=\linewidth]{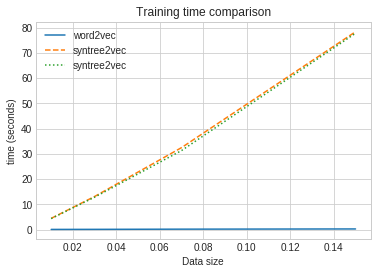}
  \caption{Training time comparison}
  \label{fig:train_time}
\end{figure}

Our syntree2vec algorithm is compared to word2vec and node2vec model in terms of training time. Their details are presented in Figure \ref{fig:train_time}. We then move towards the usability side of the algorithm and compare their transfer learning capacity on a simple supervised task. The details of the metrics and procedure is provided in the next section.

\subsection{Procedure}
We first generate the embeddings for all the models. We then report the perplexity and present some example sentences generated from the LSTM(generator) model at the end of training for varying data sizes.

\subsection{Evaluation Metrics}
The evaluation metrics  are chosen as per \cite{kawthekarevaluating}
\begin{enumerate}

    \item Perplexity\newline
    Perplexity refers to the prediction capability of the language model. If the model is less perplexed then it is a good model.It is calculated as \(2^H\). Where H is the cross entropy loss
    \item Qualitative Judgment\newline
    We display the results of some sequences generated for some sample words at the end of all epochs for varied data sizes for human inspection for an insight into the quality of the model.
    \footnote{\label{down}https://github.com/shubham0704/syntree2vec}
\end{enumerate}

\subsection{Results}

\begin{table}[ht]
\caption{Perplexity Scores}
\begin{center}
\scalebox{0.85}{
\begin{tabular}{ |c|c|c|c| } 
\hline
No. of sentences & node2vec & syntree2vec & word2vec \\
  \hline
  0.01MB : 73 & 24.93 & 20.11 & 22 \\ 
  0.03MB : 220 & 28.44 & 28.24 & 28.44 \\ 
  0.07MB : 508 & 43.59 & 43.41 & 43.41 \\ 
  0.15MB : 1070 & 71.20 & 71.01 & 71.01 \\

  \hline
\end{tabular}}
\end{center}
\end{table}
Since the loss, perplexity of syntree2vec is lower than word2vec, node2vec over most of the data sizes given below we say that the syntree2vec performs slightly better than both of them. Scaling this on more data would help determine the clear winner. We see that adaption of our strategy of focusing the node2vec has helped improve the learning capacity of the model. There is a clear margin of difference between the perplexity scores of node2vec and syntree2vec.
The sentences generated for these data sizes on each epoch are provided online. Refer \ref{down}. There is a tradeoff in training time about which the readers must beware. Our training time is significantly higher than the word2vec model. We therefore recommend it to be used for small data sizes.

\section{Discussion and Conclusion}
In this paper we studied feature learning from a graph search perspective.
This perspective helped the algorithm work with a more global search strategy than it was previously constrained to. We saw that our algorithm performs slightly better than word2vec model and improves the node2vec algorithms ability to work on text data for feature learning. Further study in this direction would surely make it better. We introduced a new trend of analyzing word embeddings based on varied data sizes and tested them on a simple supervised task.

\end{document}